\documentclass[runningheads]{llncs}
\usepackage[T1]{fontenc}
\usepackage{graphicx}
\usepackage{amssymb}
\usepackage{amsmath}
\usepackage{multirow}
\usepackage{arydshln}
\usepackage{subcaption}

\begin{document}

\title{Test-Time Augmentation for Tabular-to-Image Classifiers under Distribution Shifts}

\titlerunning{Test-Time Augmentation for Tabular-to-Image Classifiers}
%

\author{
Malena Loza\inst{1} \and
Felipe Grijalva\inst{1} \and 
Eva Milara\inst{2} \and 
Luis Bote Curiel\inst{2} \and 
Francisco J. Lara Abelenda \inst{3} \and
David Chushig-Muzo\inst{2}
}

\authorrunning{Loza-Casa M. et al.}

\institute{
Colegio de Ciencias e Ingenierías, Universidad San Francisco de Quito (USFQ), Quito, Ecuador \email{\{mloza,fgrijalva\}@usfq.edu.ec} 
\and
Department of Signal Theory and Communications, Telematics and Computing Systems, Rey Juan Carlos University, Madrid, Spain \\ \email{\{eva.milara,luis.bote,david.chushig\}@urjc.es} 
\and
Faculty of Experimental Sciences, Universidad Francisco de Vitoria, Madrid, Spain \email{francisco.lara@ufv.es} 
}

\maketitle 

\begin{abstract}
Tabular-to-image methods that convert tabular data into visual representations have emerged as a novel paradigm for leveraging the high performance of deep learning models. Despite their advantages, the robustness of these methods under distribution shifts remains under explored. Test-Time Augmentation (TTA) is an effective approach in image classification to improve model generalization and robustness, where predictions over multiple transformed views of each input are aggregated. This work evaluates the impact of TTA techniques on predictive performance under Out-Of-Distribution (OOD) for representations generated by tabular-to-image methods. Six tabular-to-image encoding methods were considered: TINTO, IGTD, DeepInsight, BIE, DistanceMatrix, Fotomics. Twenty-five TTA techniques were used, organized into six types: Geometric, Photometric, Structural, Frequency/Encoding, Mixup, and Composite. We employed two datasets from the TableShift benchmark (HELOC and Voting) that provide in-distribution and OOD test subsets designed to evaluate the effect of distribution shifts on tabular data. The results indicate that TTA improves OOD performance, with composite and photometric strategies providing the best trade-off between robustness and variance. In contrast, frequency-domain transformations that alter the encoder's feature-to-intensity mapping consistently degrade performance. These findings highlight TTA as a promising approach for improving the robustness and generalization of classifiers trained on image representations derived from tabular data, particularly under distribution shifts.
\end{abstract}

\section{Introduction}

Artificial neural networks and Deep Learning (DL) have shown great capacity to extract patterns from large data, highlighting the high predictive performance of Convolutional Neural Networks (CNNs) and Vision Transformers (ViTs) in computer vision~\cite{lara2025transfer}. Despite this, their success has not yet been extended to tabular data due to the lack of inherent spatial relationships among features~\cite{alenizy2025transforming}.

A recent line of research that transforms tabular data into image-like representations has received great attention to leverage the predictive performance of CNNs and pretrained ViTs~\cite{liu2025comprehensive}. Tabular-to-image methods capture feature relationships and underlying patterns from tabular data into spatially organized visual representations, assigning each feature value to a pixel location. Among most popular methods, we find TINTO~\cite{castillo2023tinto}, IGTD~\cite{zhu2021converting}, LM-IGTD~\cite{gomez2024lm}, DeepInsight~\cite{sharma2019deepinsight}, BIE~\cite{briner2023tabular}, DistanceMatrix~\cite{sharma2022classification}, Fotomics~\cite{zandavi2023fotomics}. These methods have demonstrated competitive performance across a range of applications such as disease classification~\cite{gomez2026tabular,lara2025transfer}, indoor localization~\cite{castillo2025mimo,liu2026interpretable}, among others.

DL models are generally trained under the assumption that training and test subsets follow the same distribution. However, in real-world applications, models trained are deployed in environments where the input has shifted due to different distribution shifts such as geographic variatio and demographic differences. Under Out-Of-Distribution (OOD) conditions, models with reasonable In-Distribution (ID) performance can substantially degrade, producing unreliable predictions~\cite{liu2023need}. The TableShift study~\cite{gardner2023benchmarking} investigated the OOD effect on tabular data, providing ID and OOD subsets constructed from different domain features (\textit{e.g.,} geographic region, demographic subgroup). It confirmed that the ID/OOD performance gap is consistent across a wide range of machine learning models~\cite{gardner2023benchmarking}.

Test-time augmentation (TTA) has emerged as an effective approach to improve robustness and calibration of computer vision models at inference time~\cite{son2023efficient}. Instead of predicting once on the original input, multiple transformations are performed on an image (\textit{e.g.,} flips, rotations, crops) at the inference stage, and the resulting predictions of these image versions are aggregated and averaged to produce a final output~\cite{son2023efficient}. TTA reduces the variance of the classifier's prediction under input perturbation, which is the approach needed to mitigate the effect of distribution shifts. Given its success in image classification, TTA is a promising approach for images generated by tabular-to-image methods. However, its impact on these representations, particularly under distribution shifts, remains unexplored.

In this paper, we evaluate the effectiveness of TTA techniques on images generated by tabular-to-image methods for enhancing robustness and model generalization under OOD. Six tabular-to-image encoding methods were considered, including: Binary Image Encoding (BIE), DeepInsight, DistanceMatrix, Fotomics, IGTD and TINTO. We evaluate twenty-five TTA techniques organized into six types: Geometric (flips, rotations), Photometric (brightness, contrast, blur, sharpening), Structural (noise, crop, cutout, elastic deformation), Frequency/Encoding (histogram equalization, solarization, posterization, channel shuffling), Mixup (self-blending with transformed copies), and Composite (chained different TTA techniques). We evaluate on two datasets from the TableShift study~\cite{gardner2023benchmarking}: HELOC and Voting. To the author's knowledge, this is the first work that investigates the effectiveness of TTA on the predictive performance under OOD for images derived from tabular-to-image methods. 

\begin{itemize}
    \item We conduct an empirical evaluation of the effectiveness of TTA techniques on images derived from tabular-to-image encoding methods.

    \item We evaluate the impact of TTA techniques on OOD generalization using real-world datasets from the TableShift study.
    
    \item We extend the OOD analysis of predictive models to tabular-to-image methods, assessing the relationship between ID and OOD performance.
    
    \item We provide empirical evidence of the robustness of tabular-to-image encoding methods under distribution shifts across different domains.

\end{itemize}

\section{Materials and methods}

\subsection{Datasets}

We employ two datasets (HELOC, Voting) from the TableShift study~\cite{gardner2023benchmarking}. HELOC is a financial risk prediction dataset used to predict whether a consumer will repay a home equity line of credit. The distribution shift is characterized by label shift, whereby the distribution of the target variable differs substantially between the ID and OOD subsets. Voting is a dataset used to determine whether an individual votes. The domain split is defined by U.S. Census region, with the Southern states comprising the OOD subset and all other regions forming the ID subset. This geographic partition introduces a regional distribution shift representative of real-world deployment, in which models trained on one region are applied to populations with different political and demographic characteristics. Table~\ref{table:summary_datasets} summarizes the datasets considered. 

\begin{table}
\centering
\caption{Datasets used in this study.}
\label{table:summary_datasets}
\begin{tabular}{|l|l|l|l|l|l|l|l|}
\hline
\textbf{Dataset} & \textbf{Shift Variable} & \textbf{Features} & \textbf{Train} & \textbf{Validation} & \textbf{ID test} & \textbf{OOD test} & \textbf{Total} \\ \hline
Voting & Geographic region & 380 & 37548 & 4693 & 4694 & 23103 & 70038 \\ \cdashline{1-8}
HELOC & Label distribution & 38 & 2220 & 278 & 278 & 6914 & 9690 \\
\hline
\end{tabular}
\end{table}

\subsection{Tabular-to-image encoding methods}

We evaluate TTA on images generated by six tabular-to-image encoding methods: BIE, DeepInsight, DistanceMatrix, Fotomics, IGTD and TINTO.

BIE converts each feature value to its binary
floating-point representation, treating each value as an n-bit string, and stacks feature-wise strings vertically into a matrix interpreted as a binary image~\cite{briner2023tabular}. The resulting image captures the full numerical precision of each feature without any spatial optimization. 

DeepInsight uses dimensionality reduction methods (including kernel PCA and t-SNE) to map features into a two-dimensional image grid, yielding a fixed coordinate for each feature that is shared across all samples~\cite{sharma2019deepinsight}. A convex hull is computed around the resulting point cloud and rotated to minimize the bounding rectangle, compressing unused space before the plane is discretized into a fixed pixel grid. Feature values falling into the same pixel are aggregated. Each sample is then rendered as an image by mapping its feature values onto this shared coordinate-to-pixel layout~\cite{sharma2019deepinsight}.

DistanceMatrix, for a sample with $d$ features, constructs a square $d \times d$ matrix, where each element is computed as the difference between feature $i$ and feature $j$, and the resulting matrix is normalized to the range $[0, 1]$~\cite{sharma2022classification}. Because the matrix is generated independently per sample from that sample's own feature values, it does not require dimensionality-reduction or embedding. To improve the spatial coherence of the images, feature ordering can be optimized by minimizing a covariance-based adjacency ranking, placing correlated features closer together before the distance matrix is computed~\cite{sharma2022classification}.

Fotomics employs the Fast Fourier Transform (FFT) rather than manifold-learning techniques such as t-SNE~\cite{zandavi2023fotomics}. After log normalization, the FFT is applied to each sample, producing a complex-valued spectrum whose real and imaginary components serve as the Cartesian coordinates of each feature. The convex hull is computed over this feature space and rotated via a standard 2D rotation matrix to orient the layout for image framing. The rotated Cartesian coordinates are then discretized into pixel coordinates based on a predetermined pixel resolution, with pixel intensity at each location set to the mean of overlapping feature values, producing one RGB image. 

IGTD converts tabular data into compact image representations by optimizing an assignment of features to pixel positions~\cite{zhu2021converting}. It first ranks the pairwise Euclidean distances between features and between pixel locations in the image grid. It then iteratively swaps features to minimize the discrepancy between the two rankings. IGTD reorders features by placing those with similar distributions across samples close together and dissimilar features farther apart. Each sample is then represented as an image in which each pixel corresponds to a feature and its intensity reflects the feature value~\cite{zhu2021converting}.

TINTO applies dimensionality reduction techniques (PCA and t-SNE) to project the feature space onto a two-dimensional plane~\cite{castillo2023tinto}. Features that are close in the original feature space are mapped to neighboring pixel locations, whereas pixel intensities indicate the corresponding feature values. TINTO enhances the image representation by applying Gaussian blurring, which smooths transitions between adjacent features while adding local contextual information~\cite{castillo2023tinto}.

\subsection{Test-time augmentation techniques}

At inference time, we applied a set of image transformations to each encoded test image, generating $K$ augmented views. The classifier predicts over all views and the predictions are aggregated into a single output. We employed twenty-five TTA techniques organized into six types: geometric, photometric, structural, frequency/encoding, mixup and composite. Geometric uses spatial transformations (\textit{e.g.,} flip, rotate) that do not alter pixel values, only their arrangement. Photometric employs adjustment of pixel intensity and contrast without moving pixels. Structural adds noise or removes (shifts) image regions. Frequency/Encoding alters the pixel value distribution (histogram-level operations). Mixup blends the canonical image with a transformed copy of itself. Composite chains multiple single transformations of different TTA techniques. A summary of TTA techniques and the views generated is shown in Table~\ref{tab:tta4tab_strategies}.

\begin{table}
\caption{Test-time augmentation methods used in this study.}
\label{tab:tta4tab_strategies}
\centering
\small
\scalebox{0.85}{
\begin{tabular}{|l|l|l|c|}
\hline
\textbf{Method} & \textbf{Type} & \textbf{Transformations} & \textbf{Views ($K$)} \\ \hline
geometric & Geometric & H-flip, V-flip, Rotate $90^\circ/180^\circ/270^\circ$ & 5 \\ \hline
photometric & Photometric   & Brightness $\pm30\%$, Contrast $\pm30\%$, Gaussian blur, & 6 \\ 
            &               & Sharpening $\times2$ & 6 \\ 
conservative & Photometric & Brightness $\pm$ and Contrast $\pm$ only & 4 \\ 
aggressive & Photometric & photometric + noise $\times3$ + crop $\times2$ & 11 \\ \hline
noise & Structural & Additive Gaussian noise ($\sigma = 8$) & 1 \\ 
crop & Structural & Center crop $85\%$ + resize & 1 \\ 
cutout & Structural & Zero random $20\%$ patch & 1 \\ 
elastic & Structural & Small elastic deformation & 1 \\  \hline
equalize & Frequency & Histogram equalization & 1 \\ 
solarize & Frequency & Invert pixels above threshold 128 & 1 \\ 
posterize & Frequency & Reduce to 4-bit depth & 1 \\ 
channel\_shuffle & Frequency & Random RGB channel permutation & 1 \\ \hline
mixup & Mixup & Blend with blurred self ($\alpha = 0.5$) & 1 \\ 
mixup\_conservative & Mixup & mixup + conservative & 5 \\ 
mixup\_photometric & Mixup & mixup + photometric & 7 \\ 
mixup\_noise & Mixup & mixup + noise & 2 \\ 
mixup\_crop & Mixup & mixup + crop & 2 \\ \hline
combined & Composite & mixup + photometric + noise + crop & 9 \\ 
combined\_new & Composite & combined + all category-1 transforms & 16 \\ 
noise\_crop & Composite & noise + crop & 2 \\ 
photometric\_noise & Composite & photometric + noise & 7 \\ 
photometric\_crop & Composite & photometric + crop & 7 \\ 
full & Composite & baseline + combined & 10 \\ 
full\_new & Composite & baseline + combined\_new & 17 \\ \hline
\end{tabular}
}
\end{table}

\section{Results}

\subsection{Experimental setup}

Tabular-to-image methods used in this study were implemented using the TINTOlib library~\cite{liu2025tintolib}. TableShift provides five subsets for each dataset (HELOC and Voting): \texttt{train}, \texttt{validation}, \texttt{id\_test}, \texttt{ood\_validation}, and \texttt{ood\_test}. The OOD subsets are constructed by partitioning on a domain variable that reflects a source of shift. We report results on \texttt{id\_test} and \texttt{ood\_test}. 
 
All generated images are resized to $64 \times 64$ pixels and converted to RGB format before being passed to the image classifier. TTA techniques are applied exclusively in image space, after encoding. For each test sample, we generate $K$ augmented views per encoder by applying each transformation in the strategy's definition to the encoded image. The classifier's softmax outputs across all $K \times E$ views (where $E$ is the number of active encoders) are averaged to produce the final prediction probability vector, from which the predicted class is taken as the argmax. We use EfficientNet-B0 as the image classifier, initialized with ImageNet-pretrained weights and fine-tuned on each dataset's training images. The model uses as input the $64 \times 64 \times 3$ RGB image produced by each encoding method and outputs a softmax probability vector over classes. We used the Adam optimizer with a learning rate of $10^{-3}$, cross-entropy loss, and a batch size of 32. 

All experiments are repeated across five random seeds to assess model generalization. We report mean Area Under the ROC Curve (AUC) and standard deviation across seeds. AUC is used as the primary metric because both HELOC and Voting datasets are imbalanced. 

\subsection{Effect of tabular-to-image encoder method on out-of-distribution}

\begin{figure}[!htbp]
    \centering
  \begin{subfigure}[b]{0.99\linewidth}
    \includegraphics[width=\linewidth]{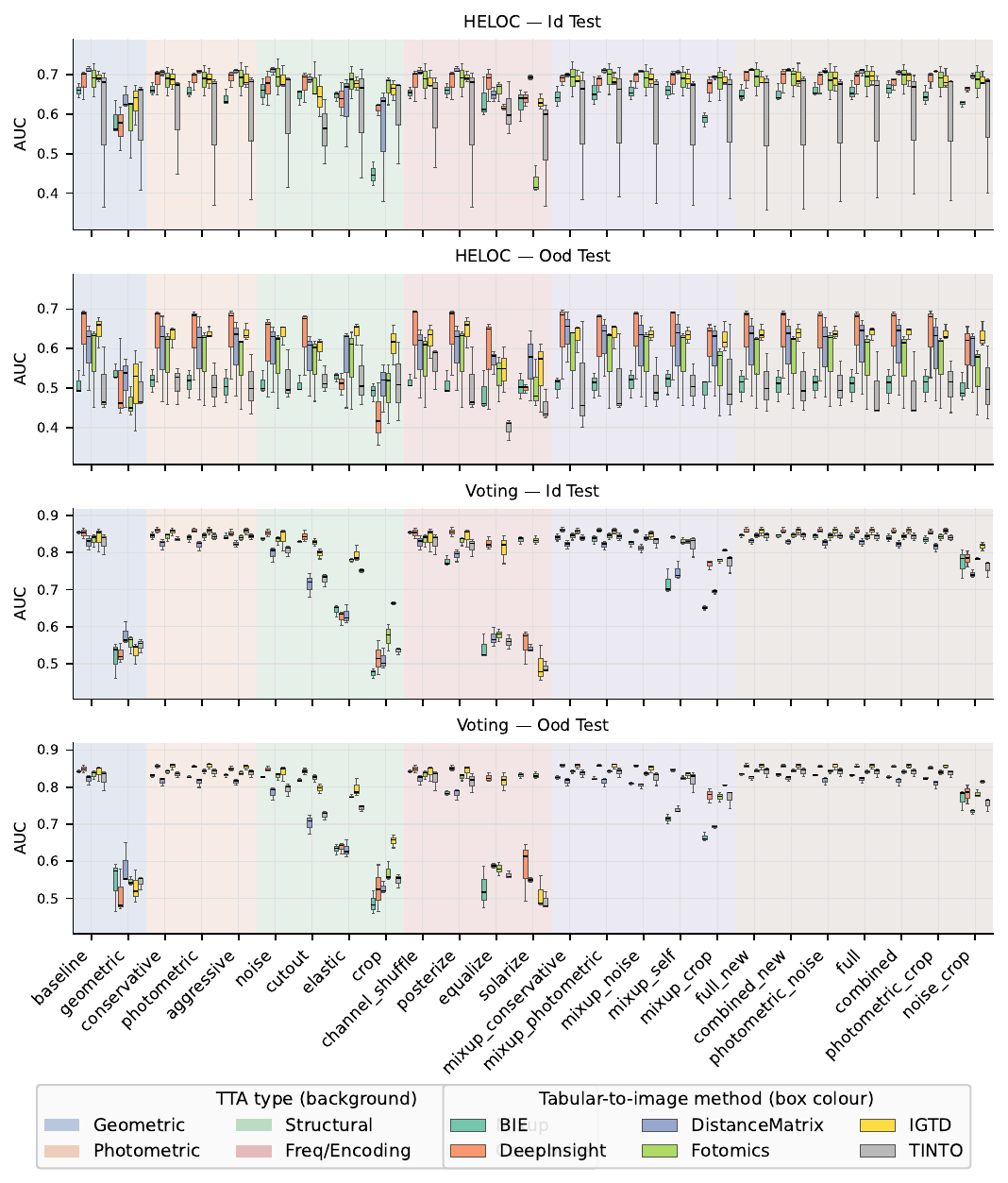} 
  \end{subfigure}
  \captionsetup{justification=justified, singlelinecheck=false, margin=6pt} 
  \caption{In-distribution and out-of-distribution AUC results for HELOC and voting datasets.}
  \label{fig:encoder_heloc_voting} 
\end{figure}

Figure~\ref{fig:encoder_heloc_voting} presents the AUC by individual TTA technique and tabular-to-image methods for HELOC and Voting datasets. On the ID (HELOC), AUC values for most TTA techniques and encoding methods are between 0.60 and 0.75, suggesting that no single encoder outperforms the others. Several TTA techniques show a decrease in AUC across most tabular-to-image methods (most notably crop, posterize, and solarize), where bars drop toward 0.45–0.55 with wide error bar. For the OOD (HELOC), the AUC distribution shifts downward by approximately 0.05–0.10 across nearly all TTA technique–encoder combinations. In addition, the performance gap between the best- and worst-performing TTA techniques becomes more pronounced. Composite and MixUp strategies consistently achieve higher median AUC values with relatively small error bars, indicating greater robustness and stability. In contrast, several structural and frequency/encoding-based strategies (e.g., Crop, CutOut, Posterize, Solarize, and Channel Shuffle) exhibit lower median AUC values. The error bars are larger under OOD test than ID test for most TTA techniques, reflecting the increased instability of model performance under distribution shift.

In contrast to HELOC, the Voting dataset yields consistently higher AUC values, with most TTA-technique–encoding-method combinations achieving scores between 0.80 and 0.90 on both subsets (ID and OOD). The error bars are tighter than HELOC results, reflecting low variance across seeds. Despite this high performance, two TTA techniques stand out as consistent exceptions across both ID and OOD: aggressive and crop in the Structural type, and channel\_shuffle and solarize in the Freq/Encoding type, all of which show sharp drops to AUC values of approximately 0.50–0.65 for several tabular-to-image methods. TINTO is the encoder most frequently exhibiting low-AUC drops, whereas DeepInsight, Fotomics, and IGTD maintain consistently high AUC even under more aggressive transformations.

\subsection{Effect of test-time augmentation types}

\begin{figure}[!htbp]
    \centering
  \begin{subfigure}[b]{0.85\linewidth}
    \includegraphics[width=\linewidth]{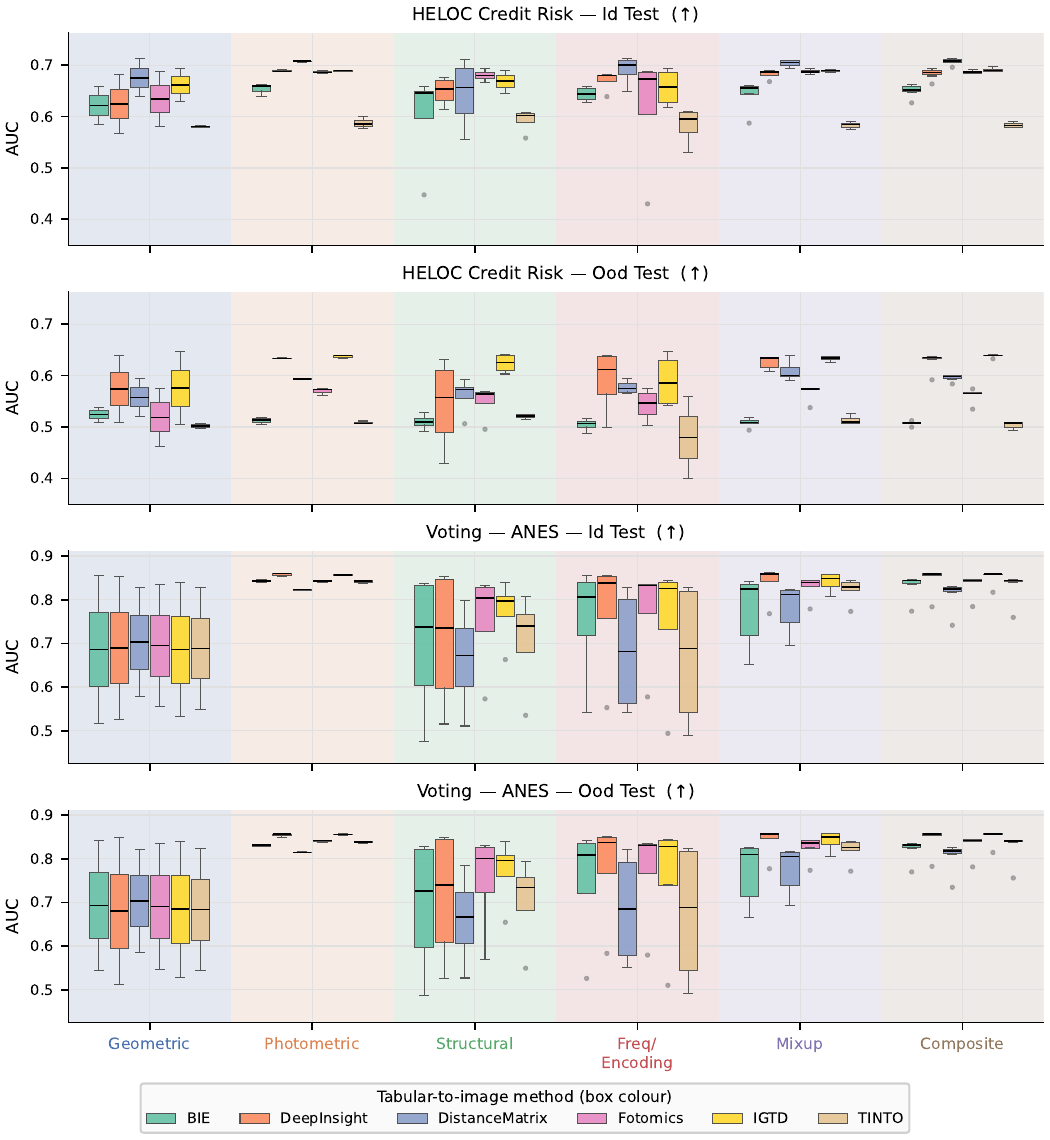} 
  \end{subfigure}
  \captionsetup{justification=justified, singlelinecheck=false, margin=6pt} 
  \caption{In-distribution and out-of-distribution AUC results for HELOC and Voting datasets grouped by TTA type. }
  \label{fig:encoder_family_heloc_voting} 
\end{figure}

Figure~\ref{fig:encoder_family_heloc_voting} shows AUC values for the HELOC and Voting datasets across the six types of TTA techniques. Each box summarizes the spread of AUC values obtained across the individual TTA approaches belonging to that type, with color indicating the tabular-to-image method used. For the ID (HELOC), AUC values are between 0.55 and 0.70, with DeepInsight, DistanceMatrix, Fotomics, and IGTD achieving higher medians than BIE and TINTO. For the OOD (HELOC), the AUC decreases by 0.05–0.10, and the performance differences between tabular-to-image encoding methods become more pronounced. DeepInsight and IGTD present high performance in the Photometric, Mixup, and Composite families, whereas TINTO presents an AUC of 0.50 in other types. 

For the Voting dataset, AUC values are tightly concentrated between approximately 0.83 and 0.86, indicating minimal variability among the individual strategies within each TTA type. This contrasts with the Geometric, Structural, and Frequency/Encoding types, which show wider interquartile ranges (roughly 0.50–0.85) and lower median AUC values, particularly for DistanceMatrix and TINTO, both of which exhibit long lower whiskers and several low outliers in the Structural and Frequency/Encoding types. Unlike the HELOC results, the Voting dataset shows minimal difference between the ID and OOD panels, suggesting that TTA on Voting is less sensitive to the distribution shift evaluated. TINTO consistently shows the lowest median AUC among encoders in the Geometric and Structural families in both splits, whereas DeepInsight, Fotomics, and IGTD maintain the most stable, high-AUC performance across nearly all types.

\section{Discussion}

This study presented a comprehensive evaluation of TTA techniques on images generated using tabular-to-image methods under distribution shifts across two TableShift datasets spanning label shift (HELOC) and geographic shift (Voting).

The most fundamental finding is that TTA does not uniformly improve robustness across all evaluation scenarios. For HELOC, where the distribution shift produces an inconsistent ID/OOD performance gap, the choice of TTA technique and encoding method has a notable effect on OOD generalization, with composite and mixup strategies retaining higher AUC and lower variance under shift compared to structural and freq/encoding techniques. For Voting, however, the ID/OOD gap is negligible across all TTA-technique-encoding method combinations. These findings suggest that the benefits of TTA for robustness are most pronounced when the distribution shift affects the classification task.

Not all image transformations are appropriate for image-derived from tabular-to-image methods. Geometric transformations (\textit{e.g.,} horizontal/vertical flips, rotations) and photometric adjustments (\textit{e.g.,} brightness, contrast) yield the most stable and consistently high AUC across both datasets and encoders. By contrast, frequency/encoding transformations (histogram equalization, solarization, posterization, and channel shuffling) produce the widest performance variance and the most severe AUC drops. In tabular-to-image encoders such as TINTO, IGTD, and DeepInsight, pixel intensity directly encodes feature magnitude, so operations that non-linearly remap the global pixel value distribution (as histogram equalization and solarization do) can disrupt the learned feature-to-intensity correspondence that the classifier relies on. Structural transformations, such as random cropping and cutout, can similarly degrade performance by removing localized feature information when feature locations are fixed by the encoder. These findings suggest that the effectiveness of a given TTA family depends on the extent to which it preserves the encoder's implicit representational relationship between feature values and pixel intensities.

Across both datasets, DeepInsight, Fotomics, and IGTD consistently achieve higher median AUC and smaller performance drops under aggressive augmentation than BIE and TINTO. In particular, TINTO shows the deepest AUC collapses under structural and frequency/encoding transformation. Composite strategies offer the best robustness-variance tradeoff. Across both datasets, the composite and mixup strategy families show the most favorable combination of high median AUC and low variance in the box plots. Strategies such as combined, combined\_new, and full\_new, which integrate photometric, mixup, and noise-based transformations while preserving geometric structure, appear to induce an implicit regularization effect. Averaging predictions over a diverse set of moderate augmentations, rather than a single aggressive transformation, makes the ensemble less susceptible to instability caused by any individual view. In contrast, single-transformation strategies from the structural and frequency/encoding categories generate only one augmented view, making the final prediction highly sensitive to the effects of a single transformation. This suggests that composite strategies employing moderate transformations and a larger number of views provide greater OOD robustness.

This study has two main limitations. First, the evaluation is restricted to two datasets from the TableShift benchmark, and extending the analysis to the remaining datasets is left for future work. Second, all experiments use a fixed EfficientNet-B0 architecture trained independently for each encoder. Joint training across encoders or a shared backbone with encoder-specific heads may improve the predictive performance.

\section{Conclusions}

This paper evaluates the effectiveness of TTA techniques on images generated by tabular-to-image methods under OOD conditions. By applying image transformations after encoding and aggregating predictions across multiple transformed views, we provide an extension of TTA to tabular data. Results from the HELOC and Voting datasets suggest that geometric and photometric transformations better preserve predictive performance because they maintain the semantic relationship between feature values and pixel intensities. In contrast, frequency- and encoding-based transformations, including histogram equalization, solarization, and posterization, alter this relationship, resulting in greater performance degradation under distribution shifts. Composite augmentation strategies combining moderate transformations from different families achieve the best balance between robustness and prediction stability. The results also indicate that the benefits of TTA are greatest under substantial distribution shift (as observed in HELOC) and more limited when the shift has little impact on classification performance (as in the Voting dataset). Our results demonstrate that TTA is an effective approach for improving robustness in classification through tabular-to-image methods. Future work will focus on extending the evaluation to additional TableShift datasets, developing encoder-aware augmentation selection strategies, and exploring adaptive prediction aggregation methods.

\begin{credits}
\subsubsection{\ackname} The authors acknowledge the Sigma AI Lab at Universidad San Francisco de Quito (USFQ) for providing the computational resources and AI infrastructure that enabled this research. Also, this work was supported through the Poli-Grants Program under Grant Numbers 41990 and 39820.

\end{credits}
%
%
%

\bibliographystyle{splncs04}
\bibliography{bibliography}

\end{document}